\documentclass{article}
\usepackage{spconf,amsmath,epsfig,amssymb,amsthm}
\usepackage{graphicx}
\usepackage{cite}
\usepackage{multirow}
\usepackage{subfigure}
\usepackage{paralist, tabularx}
\usepackage[english]{babel}

\title{Vehicle detection and counting from VHR satellite images: \\ efforts and open issues}

\name{\resizebox{\textwidth}{!}{A. Froidevaux$^{1,\dagger}$, A. Julier$^{1,\dagger}$, A. Lifschitz$^1$, M.-T. Pham$^2$, R. Dambreville$^2$, S. Lef\`evre$^2$, P. Lassalle$^3$, T.-L. Huynh$^1$ \thanks{$^{\dagger}$ Both authors contributed equally. This work is funded by the French Space Agency (CNES) under the DAJ/AR/NO-2018.009398 project.}}}
\address{
$^1$QuantCube Technology, 75002 Paris, France\\
$^2$Univ. Bretagne Sud - IRISA, 56000 Vannes, France\\
$^3$Centre National d'Etudes Spatiales, 31400 Toulouse, France \\
\texttt{aj@q3-technology.com, minh-tan.pham@irisa.fr,pierre.lassalle@cnes.fr}}

\begin{document}
%
\maketitle
\begin{abstract}
Detection of new infrastructures (commercial, logistics, industrial or residential) from satellite images constitutes a proven method to investigate and follow economic and urban growth. The level of activities or exploitation of these sites may be hardly determined by building inspection, but could be inferred from vehicle presence from nearby streets and parking lots. We present in this paper two deep learning-based models for vehicle counting from optical satellite images coming from the Pleiades sensor at 50-cm spatial resolution. Both segmentation (Tiramisu) and detection (YOLO) architectures were investigated. These networks were adapted, trained and validated on a data set including 87k vehicles, annotated using an interactive semi-automatic tool developed by the authors. Experimental results show that both segmentation and detection models could achieve a precision rate higher than 85 $\%$ with a recall rate also high (76.4 $\%$ and 71.9 $\%$ for Tiramisu and YOLO respectively).
\end{abstract}
%

\section{Introduction}
\label{sec:intro}
Vehicle detection and counting from remote sensing images is an active research topic applied to the surveillance of traffic and transportation systems, the investigation of new commercial or industrial infrastructures, as well as the study of urbanization levels, etc. Within the last decade, this task has attracted many researchers thanks to the development of very high resolution (VHR) imagery technologies together with modern object detection frameworks in the computer vision and machine learning domains. Many studies have exploited aerial images and proved their effectiveness to detect vehicles from images with spatial resolution higher than 0.3 m \cite{audebert2017segment, zhou2018robust,yu2019vehicle}. However, exploiting satellite images with spatial resolution from 0.5 m to 1 m would be more interesting since these images can cover large areas of different locations on the Earth surface over a long period of time and are more affordable for large-scale studies. In this work, we focus on the detection and counting of vehicles from VHR images acquired by the 0.5-m Pleiades satellite from different environments including urban, semi-urban and industrial zones. 

From the literature, many methods have been proposed to tackle vehicle detection using traditional pattern recognition approaches \cite{eikvil2009classification, leitloff2010vehicle, bar2013moving} as well as modern deep learning techniques \cite{chen2014vehicle,cao2016vehicle,  cao2017weakly}. Traditional approaches usually perform binary classification (vehicle and non-vehicle classes) based on local features of vehicles such as spectral profiles, gradient information (HOG, SIFT), Haar-like features \cite{eikvil2009classification, leitloff2010vehicle, bar2013moving}, etc. They mostly focus on detecting vehicles only on streets so that GIS road vector maps could be imported to limit the searching regions. Recent deep learning methods based on convolutional neural networks have achieved better performance thanks to the capacity of deep models to extract and learn vehicle characteristics within end-to-end frameworks. In \cite{chen2014vehicle}, a hybrid deep neural network was exploited to replace traditional classifiers but still based on sliding window approach.
In \cite{cao2016vehicle}, the authors proposed a transfer learning approach which helps to detect vehicles from satellite images by a detector trained on higher-resolution aerial images. However, we remark that most related studies on vehicle detection from satellite images exploit very limited data collected from Google Earth \cite{chen2014vehicle,cao2016vehicle,cao2017weakly} and rarely deal with large amount of image data coming from specific satellites. In this work, we tackle this task by exploiting recently acquired Pleiades images at 0.5-m resolution from different environments and not limited to roads. We also develop a semi-automatic annotation tool to create a large database with more than 87k vehicles. We then perform detection and counting tasks by two deep learning models: one based on a semantic segmentation framework (Tiramisu \cite{jegou2017one}) with post-processing; the other based on an adapted YOLOv3 detector \cite{redmon2018yolov3}. Both frameworks could provide good performance on our database (precision higher than $85\%$) and prove to be exploitable for large-scale traffic and urban studies.

The remainder of this paper is organized as follows. Section \ref{sec:data} describes the satellite data and the semi-automatic annotation process used to create our database. In Section \ref{sec:method}, we present the two aforementioned deep learning models for vehicle detection and counting. Section \ref{sec:results} performs qualitative and quantitative evaluations of both models. Section \ref{sec:conclusion} finally draws conclusions and discusses open issues for future work.

\section{Data creation and annotation}
\label{sec:data}
\subsection{VHR Pleiades data}
In this work, we exploit VHR images acquired by the Pleiades  satellites (PHR-1A and PHR-1B), launched by the French Space Agency (CNES), Distribution Airbus DS. Our data include 20 images of 50-cm spatial resolution acquired from region of Paris, France. These images are pan-sharpened products obtained by the fusion of 50-cm panchromatic data (70 cm at nadir, resampled at 50 cm) and 2-m multispectral images (visible RGB and infrared bands). They cover a large region of heterogeneous environments including rural, forest, residential as well as industrial areas, where the appearance of vehicles is influenced by shadow and occlusion effects. 

\subsection{Semi-automatic annotation}

\begin{figure}[!ht]
\begin{minipage}[b]{0.49\linewidth}
\centering
  \includegraphics[width=\linewidth, height = 43mm]{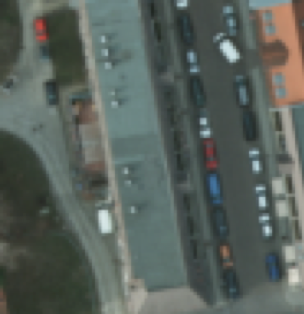} \\
  \footnotesize{(5cm aerial image downscaled to a 50cm resolution)}
\end{minipage}
\hfill
\begin{minipage}[b]{0.49\linewidth}
\centering
  \includegraphics[width=\linewidth, height = 43mm]{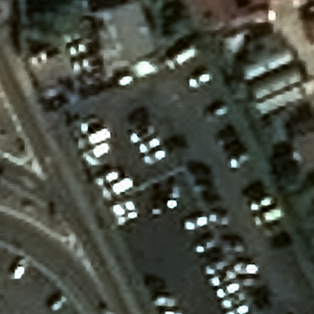} \\
  \footnotesize{(Satellite image acquired by 50cm Pleiades sensor)}
\end{minipage}
\caption{Illustration: even if the resolution of aerial/satellite images are identical, vehicles in Pleiades images are too affected by the up-scale effects.}
\label{fig:data}
\end{figure}

As we discussed earlier, there are few studies examining the problem of detection of vehicles from satellite images with a resolution close to 50 cm and there is no labeled data set available on Pleiades images. The first step of our study was the creation of a labeled data set from the collected 20 large images provided by the CNES. 
Three different strategies were considered: semi-supervised labeling using a network trained on a public aerial image dataset adapted to the resolution of Pleiades sensor; constitution of training data by integrating synthesized images; and semi-automatic annotation of Pleiades images. Visual inspection of vehicles, especially cars, on the Pleiades images clearly shows that the first two approaches are not well suited. Figure \ref{fig:data} illustrates the issues associated with car rendering on the satellite images. We observe on the left an aerial image coming from the public Postdam dataset, downscaled from 5 cm to 50 cm, and on the right a Pleiades image. Aerial images present colorful vehicles with sharp outlines whereas Pleiades images present white or black vehicles with irregular outlines. Black cars and their shadows are often indiscernible. The aspect of these small objects is greatly affected by atmospheric and optical effects, as well as the upscale effect from pansharpening process. 
A segmentation model trained on the downscaled Postdam aerial images was indeed unable to identify almost any car in the Pleiades images. Concerning the creation of a synthetic dataset, the aspect of cars obtained from 3D rendering models is similar to that of downscaled Postdam images, which is very far from that of Pleiades images. This precludes the utilisation of the synthetic approach in our case.

Therefore, a semi-automatic annotation of Pleiades images had to be performed. To facilitate this task, a interactive labeling tool was developed. It enables to label 60$\%$ of vehicles with one click using flood-fill methods adapted for this application. Prior to using the flood-fill, the local color of the road is set by clicking on a part of it. 
The flood-fill algorithm selects a connected area around a given pixel using conditions on  the distance in the HSV colorspace between the selected pixels themselves and with respect to the road pixels. HSV space is preferable to RGB because of the lack of color of both cars and roads, only the S and V components are actually used by the algorithm. 
Some vehicles require manual labeling using straight line/free hand tools included in the application. The mask color is automatically changed in order to distinguish two glued cars and extract in a second time all the different bounding boxes. 
The created data set contains 87000 annotated vehicles from different environments depending on the level of urbanization as well as the brightness and sharpness of the observed zones. This diversity of the dataset is a key point concerning the applications foreseen, as it enables our models to be reliable on various real life environments.
Data augmentation techniques were used to increase the total size of the dataset, and also to enhance its robustness. 
\section{Architecture of detection frameworks}
\label{sec:method}
Vehicle detection from satellite images is a particular case of object detection as objects are uniform, very small (around $5\times 8$ pixels/vehicle in Pleiades images) and do not overlap. To tackle this task, we investigated a segmentation algorithm Tiramisu \cite{jegou2017one} with post-processing and we adapted a direct detection network YOLOv3 \cite{redmon2018yolov3}. We now provide details about the architectures of our two models.
\subsection{Tiramisu model}
The Tiramisu model \cite{jegou2017one} with DenseNet-56 layers was implemented and data augmentation was added to the classical architecture at the predicting phase with padding and geometric operations. Consequently to this augmentation, there are in average 20 predictions per pixel that are processed by a voting system.
For later applications, we developed an indicator of the number of cars in an image.  An estimation of the number of vehicles is computed based on the size and shape of the predictions within the segmentation mask. The number of pixels associated to a block of multiple cars is divided by the mean number of pixels observed either in lined cars or in side by side cars. 

\subsection{YOLO model}
One-stage object detector YOLO (You only look once) is currently one of the most powerful models in the literature to detect visual objects within many real-time applications. The YOLOv3 model \cite{redmon2018yolov3} with Darknet-53 base network and three detection levels has been proved to be able to detect objects at different sizes, thus small objects from general computer vision tasks. However, our experiments showed that it failed to detect vehicles of approximately $5\times 8$ pixels from satellite Pleiades images. To deal with this challenge, we adapted this model by removing two detection levels related to large objects (with a sub-sampling factor called stride of 32 and 16) that we are not interested in and replacing them with two new prediction levels with strides of 4 and 2. This allows our YOLO model to work on finer cells in order to search and detect very small objects in the image. Also, new anchor sizes were computed from our training data for parameter setting, which allows the network to learn spatial shapes of annotated vehicles from the data set. Regarding the implementation, the input image size was set to $512\times 512$ pixels and the model mostly converged after 10k training iterations with a learning rate of $10^{-3}$ with decay of $5\times 10^{-4}$ (based on this code\footnote{https://github.com/AlexeyAB/darknet}).
\section{Experimental results}
\label{sec:results}
Experimental results show that Pleiades satellite images are exploitable for vehicle detection and counting applications. Performances obtained on several images are quite similar, which validates consistency of the predictions. For quantitative evaluation, we compute the recall, precision and number of vehicles counted from each model. The results presented in Table \ref{tab:results} are based on a validation set of 2673 vehicles belonging to an image including both urban and industrial zones, where vehicles appear not only on streets but on parking lots as well. As observed from the table, Tiramisu leads in terms of recall and precision rates, but YOLO also provides a reliable counting system. Model comparison is not trivial since Tiramisu is a segmentation model which requires post-processing to count vehicles while YOLO is a detection model which can directly provide the number of detected objects. Therefore both strategies have their benefits.

\begin{table}[!htbp]
\centering
\begin{tabular}{c|c|c}
 & {\bf Tiramisu} & {\bf YOLO} \\
 \hline
 Counted vehicles & 2334 & 2258 \\
TP & 2042 & 1922 \\
 FP & 325 & 336 \\
 FN & 631 & 751 \\
 \hline
 {\bf Recall} & 76.4\% & 71.9 \%\\
 {\bf Precision} & 86.2 \% & 85.1\% \\
\end{tabular}
\caption{Detection results yielded by Tiramisu and YOLO models from a validation set including 2673 vehicles.}
\label{tab:results}
\end{table}

\begin{figure*}[!ht]
\centering
\begin{minipage}[b]{0.95\linewidth}
\centering
  \includegraphics[width=145mm]{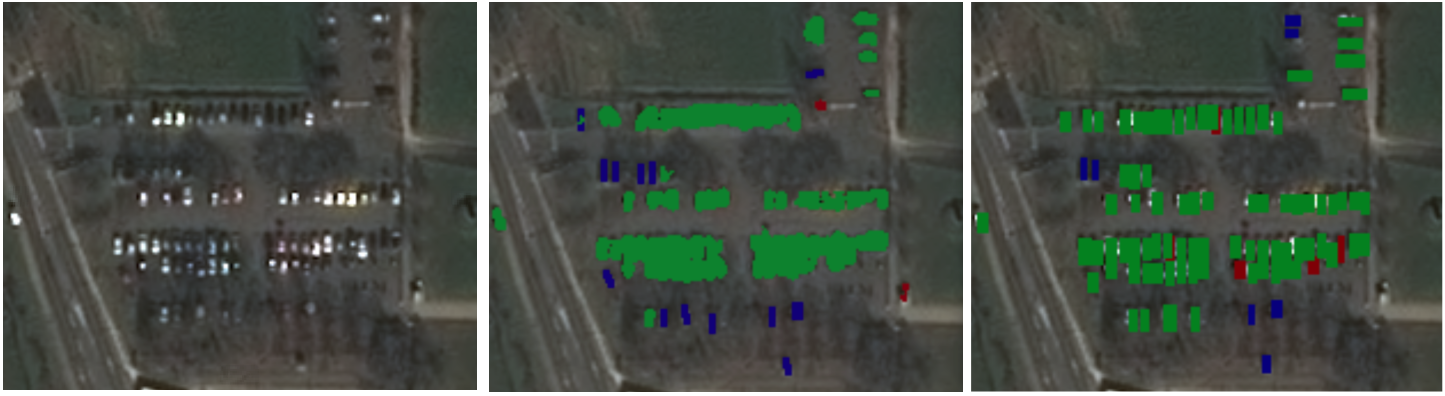}  \\
  \footnotesize{(a)}
\end{minipage}
\vfill
\vspace{1mm}
\begin{minipage}[b]{0.95\linewidth}
\centering
  \includegraphics[width=145mm]{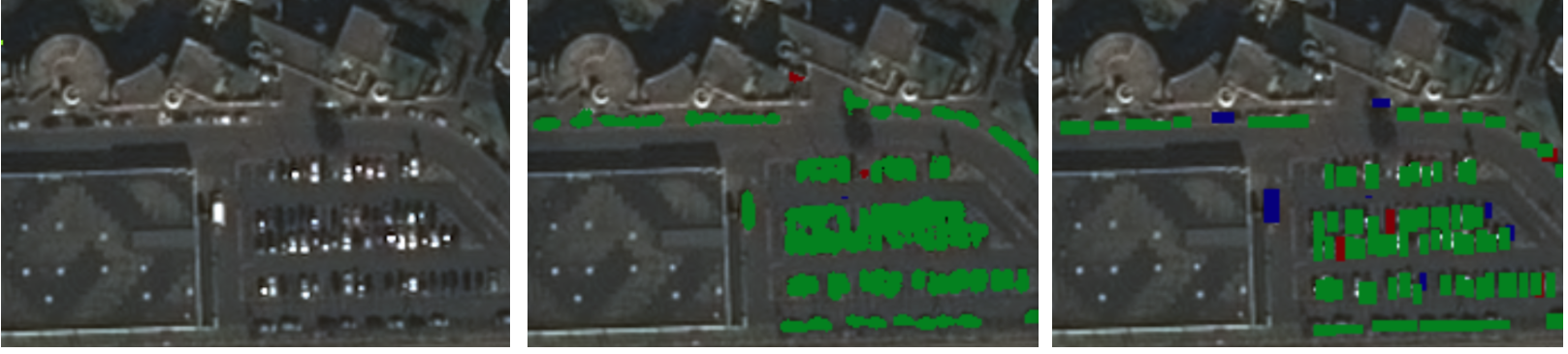}  \\
  \footnotesize{(b)}
\end{minipage}
\caption{Detection results using Tiramisu (middle) and YOLO model (right). Color codes: green for good detection, blue for missed detection and red for false alarms.}
\label{fig:results}
\end{figure*}

We show in Figure \ref{fig:results} some illustrations of detection results from two selected zones including streets and parking lots. As we can see from the figure, both models are able to detect cars on parking lots as well as on streets, while previous literature research mostly focused on streets \cite{leitloff2010vehicle, bar2013moving,cao2016vehicle,yu2019vehicle}. Undetected cars are usually the ones that are difficult to identify even for a human eyes, as we can see at the bottom of Figure \ref{fig:results}-a. In some cases, it is unclear whether the model predicted a false positive or not, since the human labeling might include a few omissions due to the lack of visibility. Another remark is that false positives in YOLO predictions often come from the fact that two detected bounding boxes were predicted for a single vehicle, and only one of them can be considered as correct. 

Comparing the results of each model, we notice that there are vehicles detected by YOLO but not by Tiramisu and vice versa. Therefore a mixed model based on both predictions may increase the number of detected vehicles. We have developed a first version of a mixed model whose predictions are the ensemble formed by the union of cars detected by YOLO with a high confidence score and cars detected by both models using a lower threshold for YOLO's confidence score. 
As expected, the obtained predictions show a slight improvement of the recall rate while precision is slightly lower. On the same validation set, recall is $80.3\%$ and precision is $81.8\%$. Another interesting aspect of this mixed model is that it provides a better counting of the number of vehicles since recall and precision are more balanced : this mixed model detects 2623 vehicles on the validation set while the labeling indicates 2673 vehicles. 

\section{Conclusion and opened issues}
\label{sec:conclusion}

Both segmentation and detection models investigated in this work have proved to be able to achieve good performance for vehicle detection and counting from satellite images. Even the Tiramisu model has showed superior recall/precision rates on the studied data set, the performance of an adapted YOLOv3 model, a one-stage object detector which is not designed to detect small objects, is particularly remarkable. We note that the rendering quality of vehicles, especially dark ones on some areas of the images makes them hardly discernible to the human eyes, and small errors on the validation set can not be excluded. Because of this, the precision of both models could actually be slightly better. For the resolution available in commercial satellites (up to 30 cm), cars are arguably the smallest objects which can be detected. At this spatial scale, the aspect of objects strongly depends on the satellite optics and post-processing of the raw images. Because of this, the application of the trained models to images coming from other satellites than Pleiades, even when keeping the resolution, constitutes an open issue. 

Current results are good enough for some applications expected using this technology, such as the detection of ghost neighborhoods and industrial sites. Its application for tracking the activity levels in commercial and industrial zones could require a better quantification of the estimated accuracy dedicated to a given environment and image's acquisition conditions, to avoid biases when comparing images taken at different times or regions.

\begin{small}
\bibliographystyle{ieeetr}
\bibliography{biblio}

\begin{thebibliography}{10}

\bibitem{audebert2017segment}
N.~Audebert, B.~Le~Saux, and S.~Lef{\`e}vre, ``Segment-before-detect: Vehicle
  detection and classification through semantic segmentation of aerial
  images,'' {\em Remote Sens.}, vol.~9, no.~4, p.~368, 2017.

\bibitem{zhou2018robust}
H.~Zhou, L.~Wei, C.~P. Lim, S.~Nahavandi, {\em et~al.}, ``Robust vehicle
  detection in aerial images using bag-of-words and orientation aware
  scanning,'' {\em IEEE Trans. Geosci. Remote Sens.}, vol.~56, no.~12,
  pp.~7074--7085, 2018.

\bibitem{yu2019vehicle}
Y.~Yu, T.~Gu, H.~Guan, D.~Li, and S.~Jin, ``Vehicle detection from
  high-resolution remote sensing imagery using convolutional capsule
  networks,'' {\em IEEE Geosci. Remote Sens. Lett.}, 2019.

\bibitem{eikvil2009classification}
L.~Eikvil, L.~Aurdal, and H.~Koren, ``Classification-based vehicle detection in
  high-resolution satellite images,'' {\em ISPRS J. Photogram. Remote Sens.},
  vol.~64, no.~1, pp.~65--72, 2009.

\bibitem{leitloff2010vehicle}
J.~Leitloff, S.~Hinz, and U.~Stilla, ``Vehicle detection in very high
  resolution satellite images of city areas,'' {\em IEEE Trans. Geosci. Remote
  Sens.}, vol.~48, no.~7, pp.~2795--2806, 2010.

\bibitem{bar2013moving}
D.~E. Bar and S.~Raboy, ``Moving car detection and spectral restoration in a
  single satellite worldview-2 imagery,'' {\em IEEE J. Sel.d Topics Appl. Earth
  Observ. Remote Sens.}, vol.~6, no.~5, pp.~2077--2087, 2013.

\bibitem{chen2014vehicle}
X.~Chen, S.~Xiang, C.-L. Liu, and C.-H. Pan, ``Vehicle detection in satellite
  images by hybrid deep convolutional neural networks,'' {\em IEEE Geosci.
  Remote Sens. Lett.}, vol.~11, no.~10, pp.~1797--1801, 2014.

\bibitem{cao2016vehicle}
L.~Cao, C.~Wang, and J.~Li, ``Vehicle detection from highway satellite images
  via transfer learning,'' {\em Information sciences}, vol.~366, pp.~177--187,
  2016.

\bibitem{cao2017weakly}
L.~Cao, F.~Luo, L.~Chen, Y.~Sheng, H.~Wang, C.~Wang, and R.~Ji, ``Weakly
  supervised vehicle detection in satellite images via multi-instance
  discriminative learning,'' {\em Pattern Recogn.}, vol.~64, pp.~417--424,
  2017.

\bibitem{jegou2017one}
S.~J{\'e}gou, M.~Drozdzal, D.~Vazquez, A.~Romero, and Y.~Bengio, ``The one
  hundred layers tiramisu: Fully convolutional densenets for semantic
  segmentation,'' in {\em Proceedings of the IEEE CVPR Workshops}, pp.~11--19,
  2017.

\bibitem{redmon2018yolov3}
J.~Redmon and A.~Farhadi, ``Yolov3: An incremental improvement,'' {\em arXiv
  preprint arXiv:1804.02767}, 2018.

\end{thebibliography}
\end{small}

\end{document}